\documentclass[letterpaper, 10 pt, conference]{ieeeconf}

\usepackage{amssymb}
\usepackage{amsmath}
\usepackage{graphicx}
\usepackage{cite}
\usepackage{multirow}
\usepackage{color}

\IEEEoverridecommandlockouts                              

\newcommand{\hatt}{\hat{\theta}}
\newcommand{\nn}{\nonumber}

\title{\LARGE \bf
Reliable Intersection Control in Non-cooperative Environments*
}

\author{Muhammed O. Sayin$^{1,2}$, Chung-Wei Lin$^{2}$, Shinichi Shiraishi$^{2}$, and Tamer Ba\c{s}ar$^{1}$
\thanks{*This work was not supported by any organization.}
\thanks{$^{1}$University of Illinois at Urbana-Champaign, Urbana, IL 61801
        {\tt\small \{sayin2,basar1\}@illinois.edu}}%
\thanks{$^{2}$Toyota InfoTechnology Center, Mountain View, CA 94043
        {\tt\small \{msayin,cwlin,sshiraishi\}@us.toyota-itc.com}}%
}

\begin{document}

\maketitle
\thispagestyle{empty}
\pagestyle{empty}

\begin{abstract}
We propose a reliable intersection control mechanism for strategic autonomous and connected vehicles (agents) in non-cooperative environments. Each agent has access to his/her earliest possible and desired passing times, and reports a passing time to the intersection manager, who allocates the intersection temporally to the agents in a First-Come-First-Serve basis. However, the agents might have conflicting interests and can take actions strategically. To this end, we analyze the strategic behaviors of the agents and formulate Nash equilibria for all possible scenarios. Furthermore, among all Nash equilibria we identify a socially optimal equilibrium that leads to a fair intersection allocation, and correspondingly we describe a strategy-proof intersection mechanism, which achieves reliable intersection control such that the strategic agents do not have any incentive to misreport their passing times strategically.
\end{abstract}

\section{INTRODUCTION}


Instead of classical yet inefficient traffic lighting systems, a First-Come-First-Serve (FCFS) based autonomous intersection control, which utilizes the connectivity of the autonomous agents with each other and the infrastructure, e.g., an intersection manager, has been introduced in \cite{ref:Dresner08}. Each agent requests the usage of the intersection at a certain time slot, he/she has desired, and the intersection manager confirms the request if it is available, i.e., not has already been allocated for other agents, or proposes a counter request. This simple approach improves the efficiency of the intersections substantially in terms of travel time of the agents. However, the requests of the agents are evaluated one by one. In order to increase the efficiency further, Reference \cite{ref:Vasirani12} proposes a combinatorial auction based approach, in which the intersection is allocated to the agent who values the most, instead of the agent who has requested the earliest. This approach leads to relatively complex algorithms to compute the intersection allocation and requires a payment system. In \cite{ref:Elhenawy15}, the authors have proposed a chicken-game \cite{ref:Han12} inspired intersection control. The proposed game includes two players, where the players aim to minimize their delay while also avoiding any collision; and the intersection manager controls their actions, i.e., swerve or not, to achieve a Nash equilibrium \cite{ref:Basar99} of the game. Recently, in \cite{ref:Sayin17}, the authors have proposed an information driven intersection management mechanism, which improves the quality of transportation by prioritizing certain vehicles based on the information reported by them and ensures truthful disclosure of that information via a {\em payment} mechanism.

In this paper, we introduce an {\em intersection game} formulation. Different from the widely known chicken game \cite{ref:Han12}, here, we consider an intersection control scenario, where two non-cooperative autonomous agents (drivers) seek to use a single intersection resource at a {\em specific} time they desire. In that respect, a desired passing time for an agent can be considered as an estimated time such that he/she can pass through the intersection with minimum loss of comfort, e.g., without acceleration or deceleration (excluding the deceleration necessary to pass through the intersection safely). However, there is a certain time that the agents will need while using the intersection resource. Therefore, there can be a conflict between their desired intersection usage.

In order to avoid conflicts which might lead to accidents, the intersection manager provides a  FCFS based resource allocation protocol so that the agent who desires to pass through the intersection earlier will pass earlier, and if both desire to pass at the same time, then both have the same chance to pass first. To this end, the agents report their desired passing times to the intersection manager and the manager allocates the resources accordingly \cite{ref:Dresner08}. We note that, here, in order to increase the efficiency of the intersection usage, the request of {\em two} agents are evaluated together, which is different from the proposed approach in \cite{ref:Dresner08}, where the agents' requests are evaluated one by one. However, while the evaluations of more than one request can increase the efficiency, in a non-cooperative environment, the agents can also report strategically, e.g., they may not reveal their {\em true} desired passing times, in order to minimize the deviation of the times that the manager allows them to use the intersection from their desired passing times. Hence, we aim to formulate the strategic behavior of the agents in this non-cooperative environment. Furthermore, instead of {\em any} Nash equilibrium as in \cite{ref:Elhenawy15}, we seek to select the socially optimal one among multiple equilibria, while designing the intersection control mechanism.

The main contributions of the paper are as follows:
\begin{itemize}
\item We model the strategic behavior of the non-cooperative agents in FCFS based intersections and formulate the corresponding equilibrium points analytically.
\item We provide socially optimal intersection usage allocations and compute the socially optimal equilibrium strategies for the agents.
\item We propose a reliable intersection control mechanism, which ensures that the strategic agents reveal their private information truthfully, i.e., cannot rig the intersection control mechanism by misreporting their private information strategically.
\end{itemize}

The rest of this paper is organized as follows: In Section \ref{sec:prob}, we formulate the problem for two strategic autonomous agents and a single intersection resource. In Section \ref{sec:equilibrium}, we analyze the equilibrium scenarios in a strategic environment. We provide a strategy-proof intersection mechanism, where the agents reveal their private information truthfully, in Section \ref{sec:mechanism}. We conclude the paper with several remarks in Section \ref{sec:conclusion}. Appendices include proofs of technical results.

\section{PROBLEM FORMULATION}\label{sec:prob}

Consider two non-cooperative autonomous agents, agent-$1$ and agent-$2$, seeking to pass through a single intersection, which is equipped with a roadside unit that has communication radios. In close proximity of the intersection, agents, equipped with communication radios, can request the temporal intersection usage from the roadside unit, which reserves the intersection temporally to the agents according to a FCFS protocol and certain safety constraints. As an example of safety constraints, the roadside unit considers the necessary time for the agents to pass through the intersection while scheduling the intersection usage.

Each agent-$i$, for $i=1,2$, has access to private information: $\theta_{e,i},\theta_{d,i} \in \Theta$, denoting the earliest possible and the desired passing times through the intersection, respectively, where $\Theta \subset \mathbb{R}$ is a totally ordered finite set such that for any consecutive elements $\theta, \theta' \in \Theta$ we have $|\theta-\theta'| = \Delta$ and for all $\theta \in \Theta$, $\underline{\theta} \leq \theta \leq \bar{\theta}$, where $\underline{\theta}$ and $\bar{\theta}$ are specified upper and lower bounds, respectively, with, e.g., $\underline{\theta} = 0$. $\Delta$ can be viewed as the highest precision for time reporting. Each agent needs certain amount of time while passing through the intersection, which is denoted by $\delta t \in \Theta$, and we assume that $\delta t/2 \in \Theta$.

Agents report their passing times and the manager controls the intersection usage according to FCFS protocol. In particular, if agents report $\hat{\theta}_1 > \hat{\theta}_2$, then the allocated times would be given by $t_2 = \hat{\theta}_2$, since agent-$2$ comes to intersection first, and $t_1 = \max\{\hat{\theta}_1, t_2 + \delta t + \Delta\}$. Furthermore, in case of equality, i.e., $\hat{\theta}_1 = \hat{\theta}_2$, the manager allocates the intersection to one of them first randomly, e.g., by flipping a fair coin, for the sake of fairness. Therefore, for given reported times $(\hat{\theta}_1,\hat{\theta}_2)$, the protocol computes the corresponding allocated times, i.e., $(t_1,t_2) = \mathrm{FCFS}(\hat{\theta}_1,\hat{\theta}_2)$, uniquely.

Note that if the agent, who has the right to pass first, starts to use the resource after his/her allocated time, the time he/she will need to pass through the intersection might violate the allocated time for the other agent. Or if that agent does not use the intersection before the time allocated for the other agent, then the agent who has the right to pass second cannot start to use the intersection at his/her allocated time due to a security precaution. Therefore, without certain regulations, the agents would reveal practically unhelpful information. As an example, in the case of conflicting interests, if the agent with later desired passing time reports that his/her desired passing time is just {\em now}, then the manager would allocate the intersection to that agent to use starting from now unless the other agent has also reported to pass now. In order to avoid such cases, we consider that there is a regulation that the agents must use the intersection exactly at the allocated time. Since the allocated times are computed based on the reported desired passing times, it is the agents' responsibility to ensure that they will pass through the intersection as early as their reported times. However, even though this regulation can prevent the aforementioned cases, as we show later, this is not sufficient to incentivize the agents to reveal their desired passing times truthfully.

We consider that the agents are identical and have payoff functions penalizing the deviation of the allocated times $t_i$ from the desired passing times $\theta_{d,i}$:
\begin{equation}\label{eq:payoff}
u_i = c(|t_i - \theta_{d,i}|),
\end{equation}
where $c:\mathbb{R}\rightarrow\mathbb{R}$ is a strictly increasing, strictly convex function on $[0,\infty)$, e.g., $c(x) = x^2$. We consider Nash equilibrium \cite{ref:Basar99}, in which agents do not have any incentive to change their actions unilaterally, and an agent has an incentive to change his/her action if he/she can have a reduced payoff. Therefore, an action pair $(\hatt_1' ,\hatt_2')$ leads to an equilibrium provided that
\begin{align}
\hatt_1' \in &\arg\min_{\hatt_1 \in \Theta} c(|t_1 - \theta_{d,1}|),\nn\\
\hatt_2' \in &\arg\min_{\hatt_2 \in \Theta} c(|t_2 - \theta_{d,2}|),\nn
\end{align}
where $(t_1,t_2) = \mathrm{FCFS}(\hat{\theta}_1,\hat{\theta}_2)$. In the next section, we examine these equilibrium scenarios in detail.

\section{EQUILIBRIUM SCENARIOS IN A STRATEGIC ENVIRONMENT}\label{sec:equilibrium}

Without loss of generality, we can consider that $\theta_{d,1}$ is earlier than or equal to $\theta_{d,2}$, i.e., $\theta_{d,1} \leq \theta_{d,2}$. In such cases, if both agents reveal their desired times truthfully and $\theta_{d,1}  < \theta_{d,2}$, agent-$1$ would pass through the intersection first at his/her desired time $\theta_{d,1}$ and agent-$2$ would pass at the time $\max\{\theta_{d,1}+\delta t + \Delta, \theta_{d,2}\}$, and if $\theta_{d,1} = \theta_{d,2}$ then both agents would have equal chances to pass through the intersection first. However, the agents are strategic and a strategic agent would reveal the information truthfully if it is the best strategy according to his/her objective, e.g., minimizing the payoff function \eqref{eq:payoff}.

We point out that the agents do not have conflicting interests if $\theta_{d,1} + \delta t < \theta_{d,2}$. Hence in such cases, the agents would report their desired passing times truthfully and the manager would be able to allocate the intersection exactly at their reported times. Otherwise the agents have conflicting interests and in those cases, truthfulness is not an equilibrium achieving strategy for the agents in general. In that respect, the following cases lead to interesting equilibrium scenarios, e.g., even though $\theta_{d,1} < \theta_{d,2}$, agent-$2$ can incentivize agent-$1$ to report an earlier passing time than $\theta_{d,1}$ and under certain conditions, the intersection might even be allocated to agent-$2$ first. Next, we examine these scenarios in detail.

\begin{figure}[t!]
\centering
\includegraphics[width = .4\textwidth]{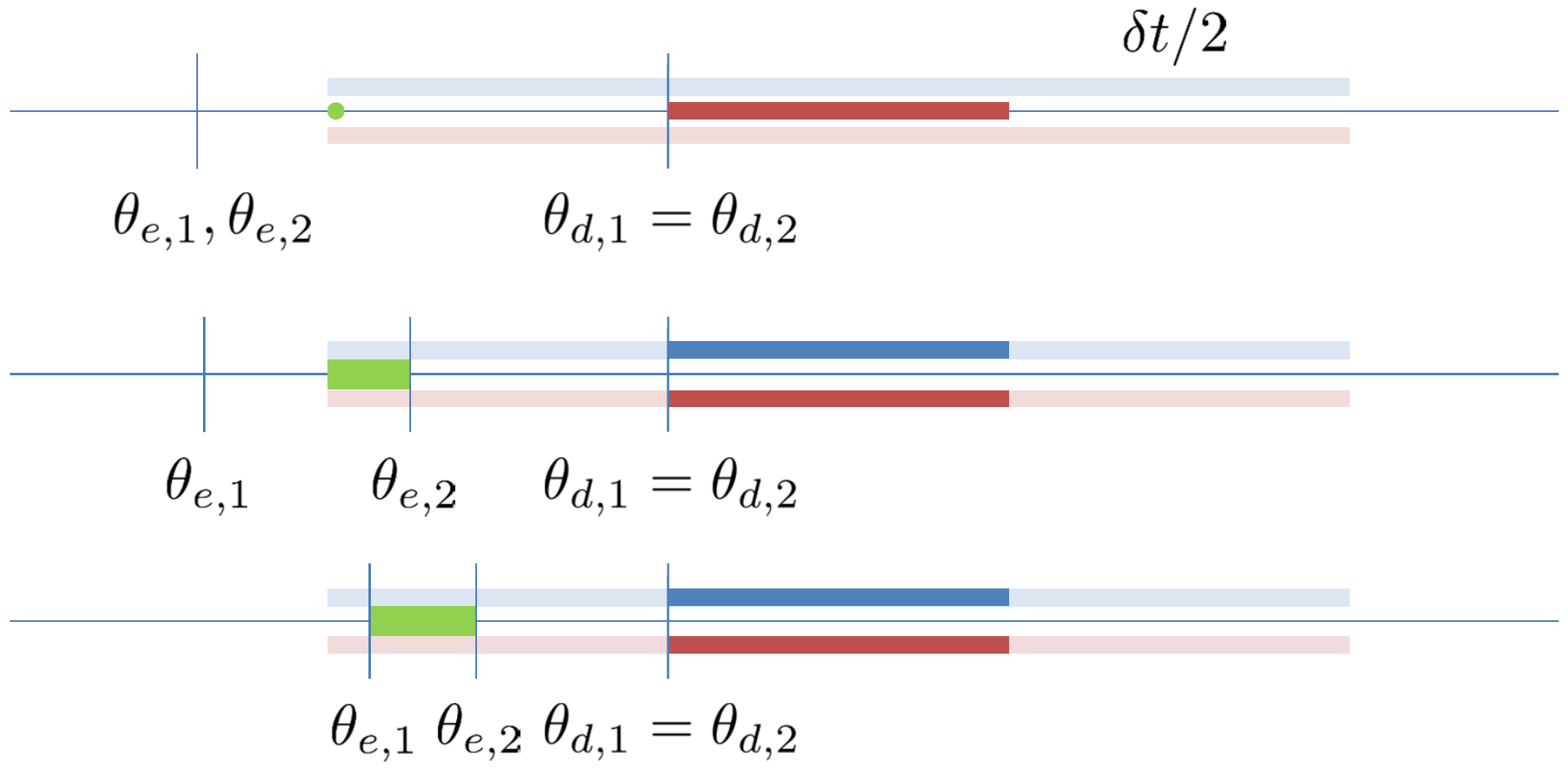}
\caption{Equilibria in the cases that $\theta_{d,1} = \theta_{d,2}$, and without loss of generality we assume $\theta_{e,1} \leq \theta_{e,2}$. The blue and red bars show a distance of $\delta t/2$ from the desired passing times. Dark and light coloring is used to represent how distant it is. As an example, right end of the bars are $\delta t$ away from $\theta_{d,i}$, $i=1,2$. Additionally, the green bars and dot show the set of equilibrium achieving actions of agent-$1$, i.e., set of $\hatt_1 \in \Theta$ such that there is a $\hatt_2 \in \Theta$ and the pair $\{\hatt_1,\hatt_2\}$ leads to an equilibrium, which are explained in Lemma 1 in detail.} \label{fig:sub1}
\end{figure}

\noindent
{\bf Lemma 1.}
{\em Let $\theta_d := \theta_{d,1} = \theta_{d,2}$. Without loss of generality, suppose $\theta_{e,1}\leq\theta_{e,2}$. Then, the action pairs $\{\hatt_1 = \theta, \hatt_2 = \theta + \Delta\}$ lead to an equilibrium if there exists $\theta \in \Theta$ such that $\theta \in M_1 := \left[\max\{\theta_{e,1},\theta_{d} - \delta t/2\}, \max\{\theta_{e,2},\theta_d - \delta t/2\}\right)$. If $M_1$ is empty, i.e.,  $\max\{\theta_{e,1},\theta_{d} - \delta t/2\} = \max\{\theta_{e,2},\theta_d - \delta t/2\}$, then the action pair $\{\hatt_1 = \theta', \hatt_2 = \theta'\}$, where $\theta' := \max\{\theta_{e,2},\theta_d - \delta t/2\}$, leads to an equilibrium.}

{\em Proof.} The proof is provided in Appendix A.  \hfill $\square$

At equilibria of certain cases, interestingly, the reported time of the agents can even be less than their earliest passing time even if there is a regulation making sure that each agent must pass through the intersection at his/her allocated time. However, since the reported time can be different from the allocated time, even though that agent knows that he/she cannot pass through the intersection that early, by reporting in that way, he/she incentivizes the other agent to report a far earlier time such that his/her allocated time will not be the reported time. We also note that the deviation of this allocated time from the desired passing time can be less than the one when he/she has reported as early as the earliest passing time. Therefore, such equilibria can be more preferable for the agent that has a later desired passing time.

\noindent
{\bf Lemma 2.}
{\em Let $\theta_{e,1}, \theta_{e,2} \leq \theta_{d,1} < \theta_{d,2} \leq \theta_{d,1} + \delta t$ and $\theta_{d,2} - \delta t \leq \theta_{d,1} < \theta_{d,2} - \delta t/2$, then the actions pairs $\{\hatt_1 = \theta, \hatt_2 = \theta+\Delta\}$ lead to an equilibrium if there exists $\theta\in\Theta$ such that $\theta \in [\max\{\theta_{d,2} - \delta t, \theta_{e,1}\}, \theta_{d,1})$. Additionally, the action pairs $\{\hatt_1 = \theta_{d,1}, \hatt_2\}$, where $\hatt_2 \in \Theta$ such that $\hatt_2 \in (\theta_{d,1},\theta_{d,1}+\delta t + \Delta]$, also lead to an equilibrium.}

\begin{figure}[t!]
\centering
\includegraphics[width = .4\textwidth]{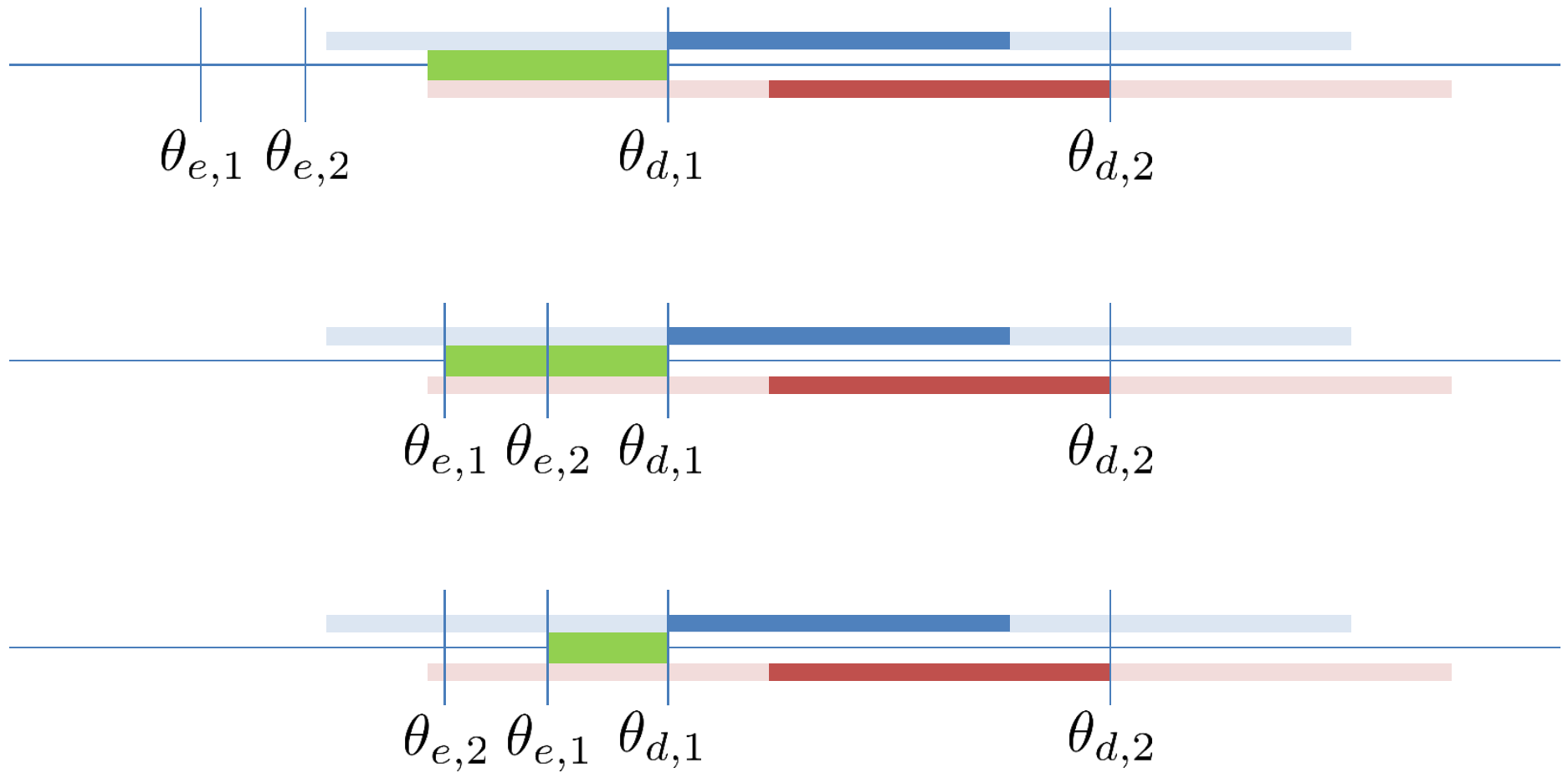}
\caption{Equilibria in the cases that $\theta_{e,1}, \theta_{e,2} \leq \theta_{d,1} < \theta_{d,2} \leq \theta_{d,1} + \delta t$ and $\theta_{d,2} - \delta t \leq \theta_{d,1} < \theta_{d,2} - \delta t/2$.} \label{fig:sub2}
\end{figure}

{\em Proof.} The proof is provided in Appendix B. \hfill $\square$

\noindent
{\bf Lemma 3.}
{\em Let $\theta_{e,1} \leq \theta_{e,2} \leq \theta_{d,1} < \theta_{d,2} \leq \theta_{d,1} + \delta t$ and $\theta_{d,1} \geq \theta_{d,2} -\delta t/2$, then the action pairs $\{\hatt_1 = \theta, \hatt_2 = \theta+\Delta\}$ lead to an equilibrium if there exists $\theta\in\Theta$ such that $\theta \in M_2:= [\max\{\theta_{e,1},\theta_{d,1}-\delta t/2\}, \max\{\theta_{d,2}-\delta t/2,\theta_{e,2}\})$. However, if $M_2$ is empty, then the action pair $\{\hatt_1 = \theta',\hatt_2 = \theta'\}$, where $\theta' := \max\{\theta_{d,2} - \delta t/2, \theta_{e,2}\}$, leads to an equilibrium. }

\begin{figure}[t!]
\centering
\includegraphics[width = .4\textwidth]{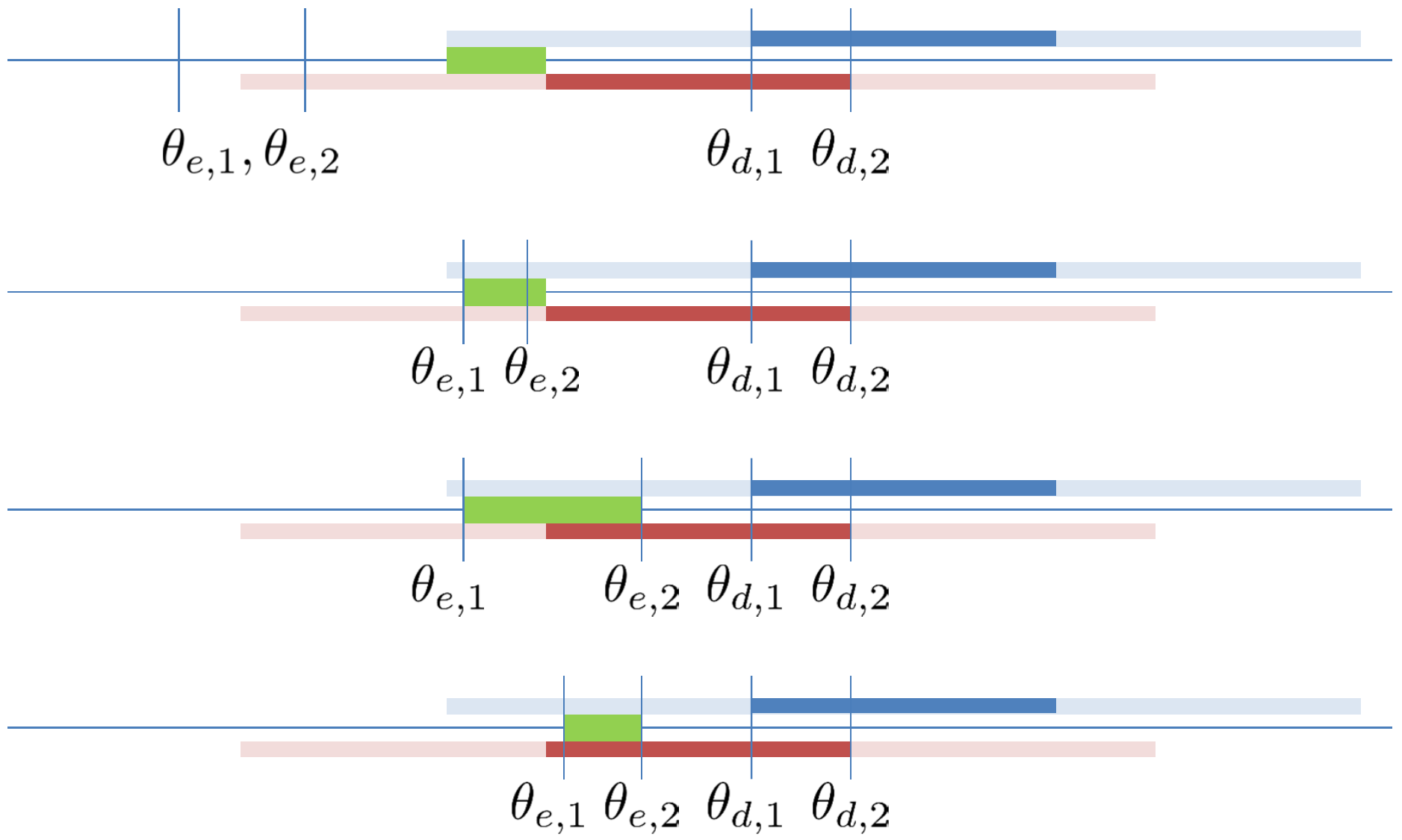}
\caption{Equilibria in the cases that $\theta_{e,1}\leq\theta_{e,2}\leq \theta_{d,1} < \theta_{d,2}$ and $\theta_{d,1} \geq \theta_{d,2}-\delta t/2$.} \label{fig:sub3}
\end{figure}

{\em Proof.} The proof is provided in Appendix C. \hfill $\square$

\noindent
{\bf Lemma 4.}
{\em Let $\theta_{e,2} < \theta_{e,1} \leq \theta_{d,1} < \theta_{d,2} \leq \theta_{d,1} + \delta t$ and $\theta_{d,1} \geq \theta_{d,2} - \delta t/2$, then we have two different cases, where the first passing agent differs. If $\theta_{e,1} \leq \theta_{d,2}-\delta t/2$, the action pairs $\{\hatt_1 = \theta, \hatt_2 = \theta+\Delta\}$ lead to an equilibrium if there exists $\theta\in\Theta$ such that $\theta \in M_3:= [\max\{\theta_{e,1},\theta_{d,1}-\delta t/2\}, \theta_{d,2}-\delta t/2)$. However, if $M_3$ is empty, then the action pair $\{\hatt_1 = \theta',\hatt_2 = \theta'\}$, where $\theta' := \theta_{d,2} - \delta t/2$, leads to an equilibrium. On the contrary, if $\theta_{d,2}-\delta t/2 < \theta_{e,1} \leq \theta_{d,1}$, the action pairs\footnote{Here, agent-$2$ passes through the intersection first at the equilibria.} $\{\hatt_1 = \theta + \Delta, \hatt_2 = \theta\}$ lead to an equilibrium if there exists $\theta\in\Theta$ such that $\theta \in M_4:= [\max\{\theta_{d,2}-\delta t/2, \theta_{e,2}\}, \theta_{e,1})$. However, if $M_4$ is empty, then the action pair $\{\hatt_1 = \theta',\hatt_2 = \theta'\}$, where $\theta' := \theta_{e,1}$, leads to an equilibrium. }

\begin{figure}[t!]
\centering
\includegraphics[width = .4\textwidth]{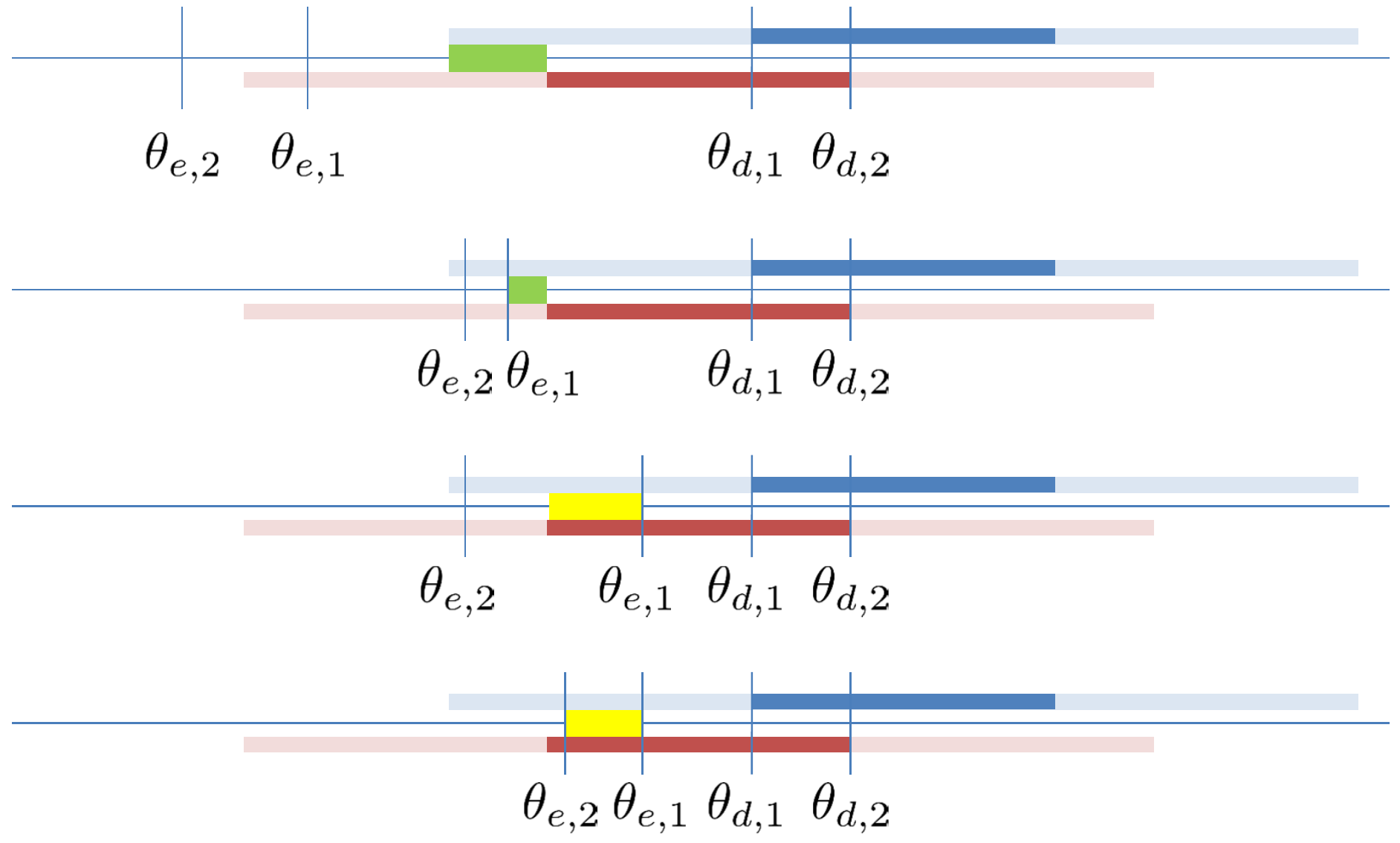}
\caption{Equilibria in the cases that $\theta_{e,2}<\theta_{e,1}\leq \theta_{d,1} < \theta_{d,2}$ and $\theta_{d,1} \geq \theta_{d,2}-\delta t/2$. Distinctively, yellow bars represent the set of equilibrium achieving actions of agent-$2$ since at those equilibria, agent-$2$ passes through the intersection first in spite of $\theta_{d,1} < \theta_{d,2}$.} \label{fig:sub4}
\end{figure}

{\em Proof.} The proof is provided in Appendix D. \hfill $\square$

\noindent
{\bf Lemma 5.}
{\em Let $\theta_{e,1} \leq \theta_{d,1} < \theta_{e,2} \leq \theta_{d,2} \leq \theta_{d,1}+\delta t$, then there exist multiple equilibria such that the equilibrium achieving action pairs are given by $\{\hatt_1 = \theta, \hatt_2 = \theta + \Delta\}$ provided that there exists $\theta \in \Theta$ such that
\begin{equation}\label{eq:sub5}
\theta \in \Big[\max\big\{\theta_{d,2}-\delta t, \theta_{d,1}-\delta t/2, \theta_{e,1}\big\}, \min\big\{\theta_{d,2}-\delta t/2,\theta_{d,1}\big\}\Big).
\end{equation}
Otherwise, there exists a unique equilibrium, where $\{\hatt_1 = \theta', \hatt_2 = \theta'\}$ and
$\theta' := \min\big\{\theta_{d,2}-\delta t/2,\theta_{d,1}\big\}$.}

\begin{figure}[t!]
\centering
\includegraphics[width = .4\textwidth]{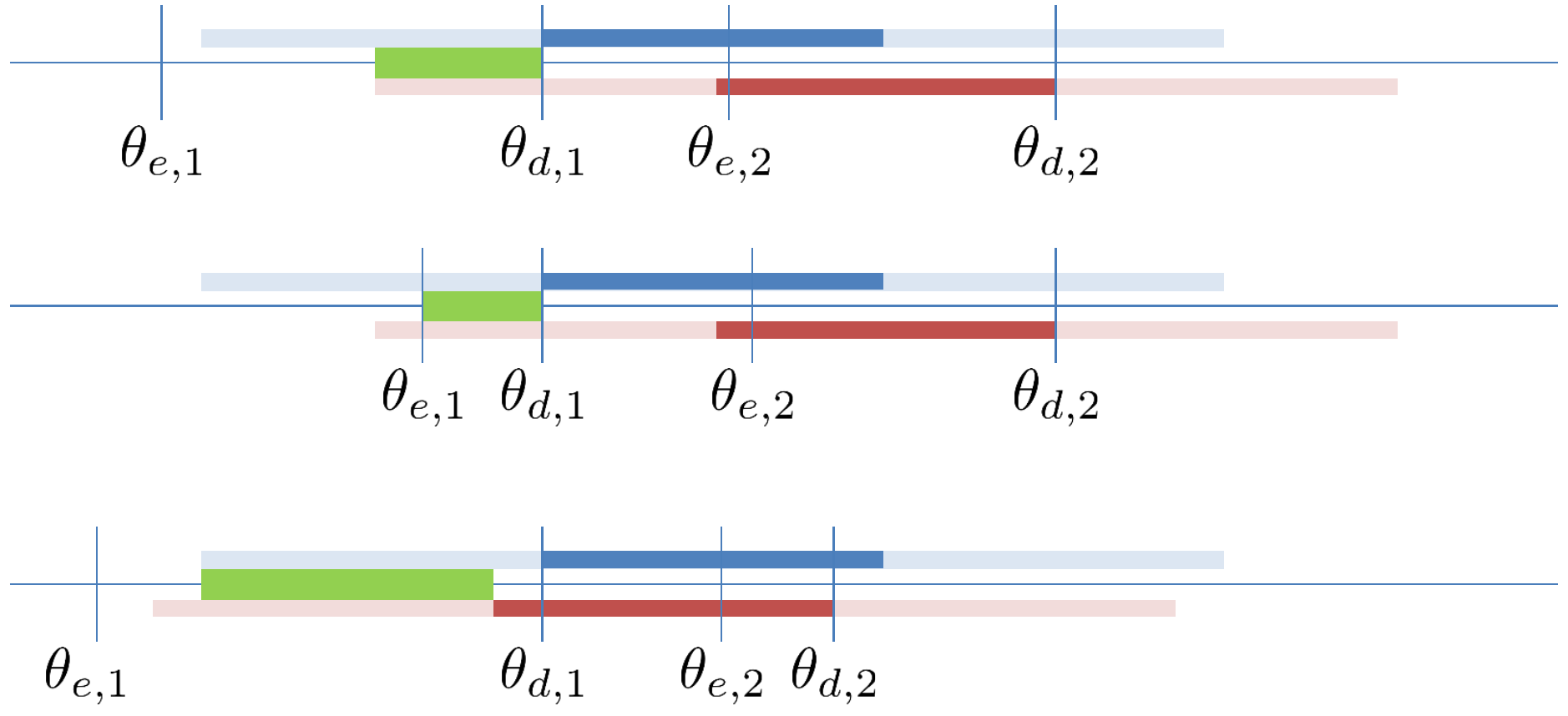}
\caption{Equilibria in the cases that $\theta_{e,1} \leq \theta_{d,1} < \theta_{e,2} \leq \theta_{d,2}\leq \theta_{d,1} + \delta t$.} \label{fig:sub5}
\end{figure}

{\em Proof.} The proof is provided in Appendix E.
\hfill $\square$

Based on Lemmas 1-5, we have the following proposition.

\begin{figure}[t!]
\centering
\includegraphics[width = .4\textwidth]{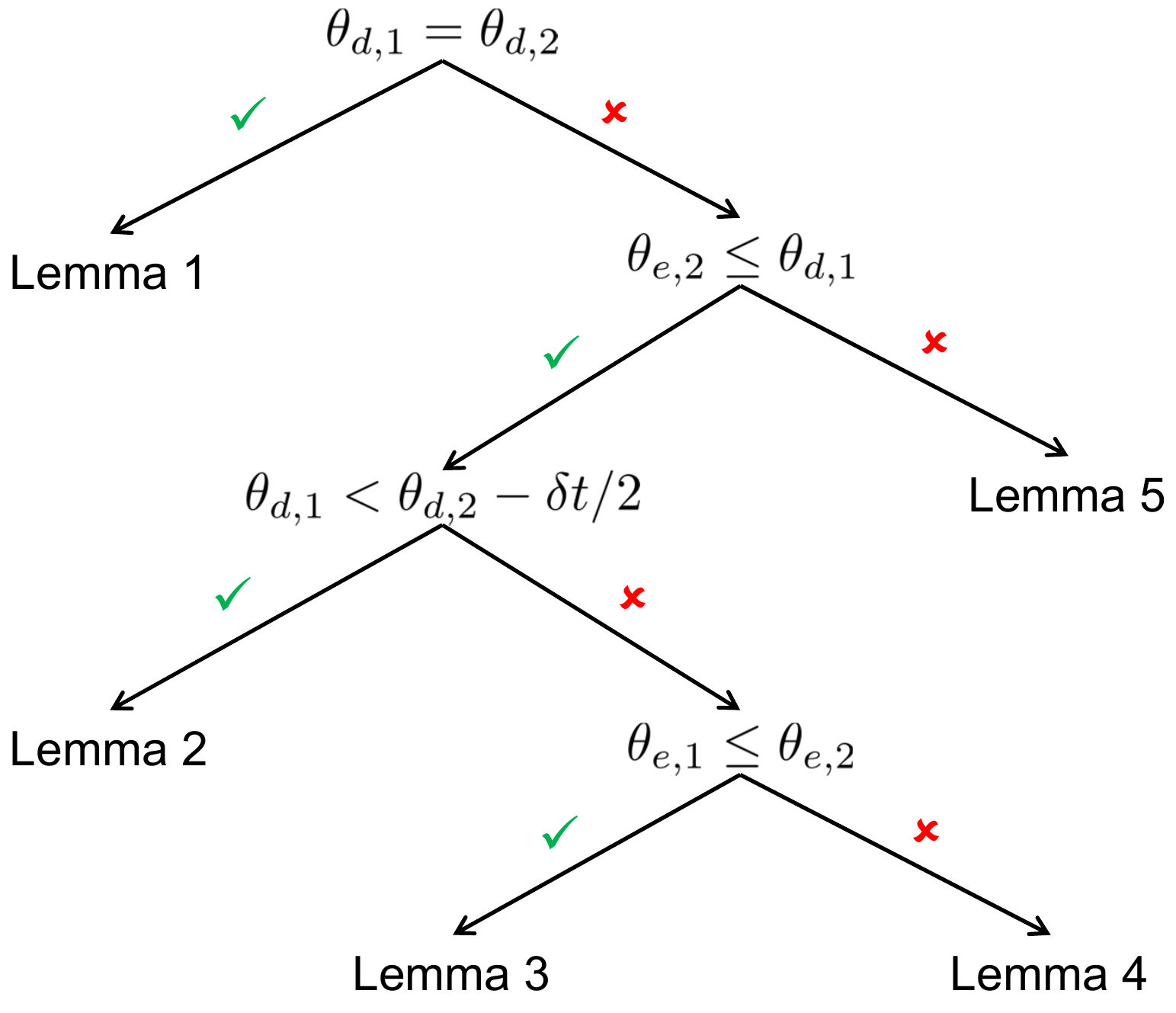}
\caption{Binary tree for the conflicting cases addressed in Lemmas 1-5 when $\theta_{d,1} \leq \theta_{d,2}$ and $\theta_{d,2} \leq \theta_{d,1} + \delta t$.}\label{fig:tree}
\end{figure}

\noindent
{\bf Proposition 1.}
{\em There exists at least one pure equilibrium point in the two-agent intersection game.}

{\em Proof.}
As seen in Fig. \ref{fig:tree}, Lemmas 1-5 cover all the conflicting cases and at each case, there exists at least one pure equilibrium. Furthermore, if there are no conflicts, then truthfulness leads to pure unique equilibrium. \hfill $\square$

We note that in general there are multiple equilibria. However, in the case of multiple equilibria, certain equilibrium points are more preferable for certain agents. Additionally, since there are multiple equilibrium points, taken actions might not yield an equilibrium. In the following section, we design a mechanism to mitigate such issues.

\section{STRATEGY-PROOF INTERSECTION MECHANISM}\label{sec:mechanism}

We seek to design an intersection mechanism between the agents and the intersection manager, where the strategic agents report their private information to the mechanism and the mechanism plays on their behalf, i.e., reports certain times to the intersection manager based on the agents' reports. Note that the agents have selfish objectives, which is to minimize their own payoff functions. However, we seek to design a mechanism which leads to minimization of the social choice function:
\begin{equation}\label{eq:social}
\sum_{i=1}^2 c(|t_i - \theta_{d,i}|).
\end{equation}
Note that our social objective \eqref{eq:social}, as a mechanism designer, can lead to different allocations from the FCFS protocol. Particularly, the socially optimal allocated times would be given by the following minimization problem:
\begin{align}
\min_{t_1,t_2 \in \Theta} &\sum_{i=1}^2 c(|t_i - \theta_{d,i}|) \mbox{ subject to} \;\;&|t_1 - t_2| \geq \delta t.\label{eq:constraint}
\end{align}
If there is no conflict of interest, i.e., $|\theta_{d,1} - \theta_{d,2}| > \delta t$, then the allocated times by the manager according to FCFS protocol also lead to the socially optimal ones. However, if there is a conflict of interest, since $c(\cdot)$ is strictly increasing and strictly convex function, the socially optimal allocated times $t_1^*, t_2^*$ are such that the deviations of the allocated times from the corresponding desired times are the same for each agent. In particular, we have\footnote{Note that we have assumed $\delta t/2 \in \Theta$.}
$i)$ if $\theta_{d,1} = \theta_{d,2}$, $t_1^* = b(\theta_{d,1} - \delta t/2) + (1-b) (\theta_{d,1} + \delta t/2 + \Delta)$, $t_2^* = (1-b)(\theta_{d,1} - \delta t/2) + b(\theta_{d,1} + \delta t/2 + \Delta))$, where $b\sim\mathrm{Ber}(1/2)$;
$ii)$ if $\theta_{d,i} < \theta_{d,j}$ and $(\theta_{d,j} - \theta_{d,i})/2 \in \Theta$, $t_i^* = \theta_{d,i} - (\delta t - (\theta_{d,j} - \theta_{d,i}))/2$, $t_j^* = \theta_{d,j} + (\delta t - (\theta_{d,j} - \theta_{d,i}))/2 + \Delta$, where $i \neq j$;
$iii)$ if $\theta_{d,i} < \theta_{d,j}$ and $(\theta_{d,j} - \theta_{d,i})/2 \notin \Theta$, $t_i^* = \theta_{d,i} - (\delta t - (\theta_{d,j} - \theta_{d,i}) + b'\,\Delta - (1-b')\Delta)/2$, $t_j^* = \theta_{d,j} + (\delta t - (\theta_{d,j} - \theta_{d,i}) + b'\,\Delta - (1-b')\Delta)/2 + \Delta$, where $b' \sim \mathrm{Ber}(1/2)$.

\noindent
{\bf Remark 1.} {\em We emphasize that the socially optimal time allocations $t_1^*,t_2^*$ do not depend on the payoff functions directly due to the assumption that agents are identical and $c(\cdot)$ is a strictly increasing and strictly convex function. Therefore, as a mechanism designer, we do not need to know the exact payoff functions except the assumption about their structure.}

\begin{figure}[t!]
\centering
\includegraphics[width = .4\textwidth]{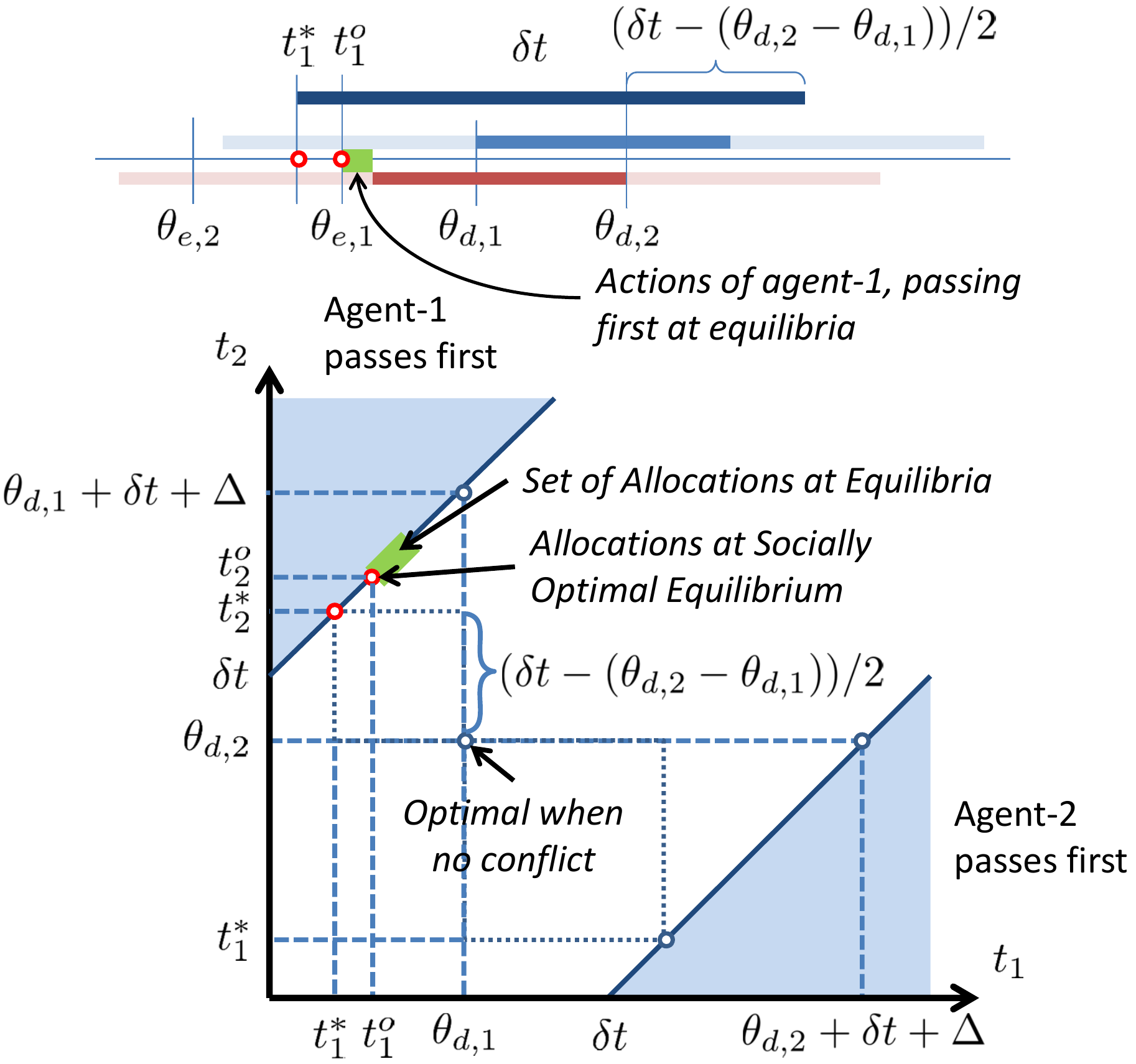}
\caption{Possible allocated times (shaded area) due to the safety constraints for the case represented at the top of the figure. Here, $\theta_{d,1} < \theta_{d,2}$ such that $(\theta_{d,2} - \theta_{d,1})/2 \in \Theta$, therefore $t_1^* = \theta_{d,1} - (\delta t - (\theta_{d,2}-\theta_{d,1}))/2$, $t_2^* = \theta_{d,2} + (\delta t - (\theta_{d,2}-\theta_{d,1}))/2$, and agent-$1$ passes first at equilibria.} \label{fig:proj1}
\end{figure}

\begin{figure}[t!]
\centering
\includegraphics[width = .4\textwidth]{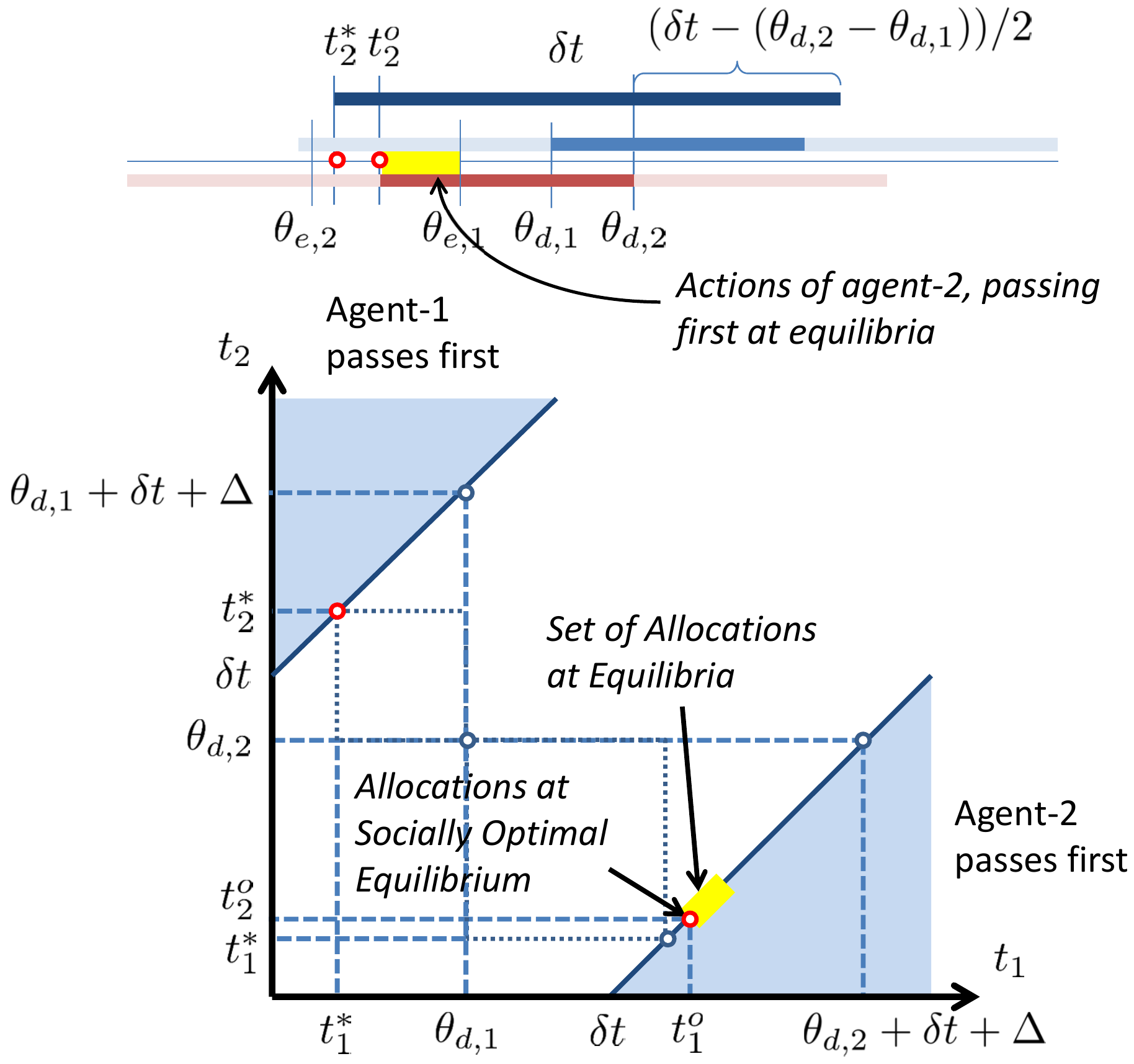}
\caption{Possible allocated times for the case represented at the top of the figure. Here, $\theta_{d,1} < \theta_{d,2}$ such that $(\theta_{d,2} - \theta_{d,1})/2 \in \Theta$, yet agent-$2$ passes first at equilibria.} \label{fig:proj2}
\end{figure}

Furthermore, in the cases of conflicting interests, the socially optimal allocated times cannot be achieved if both agents reveal their desired passing times truthfully since the manager cannot allocate a time before the reported times. If the agents behave strategically and not necessarily reveal the information truthfully, the socially optimal allocation may also not be achievable at any of the equilibria due to the earliest passing times of the agents. As an example, in Figs. \ref{fig:proj1} and \ref{fig:proj2}, we demonstrate all the possible pair of time allocations, i.e., a point in the shaded areas of the graphs, for the cases on the top of each figure. Then, the socially optimal allocation is at $t_1^* = \theta_{d,1} - (\delta t - (\theta_{d,2}-\theta_{d,1}))/2$ and $t_2^* = t_1^* + \delta t + \Delta$ while the action pair $\{\hatt_1 = \theta, \hatt_2 = \theta + \Delta\}$, where $\theta \in \Theta \cap [\theta_{e,1},\theta_{d,2}-\delta t/2)$ leads to an equilibrium in Fig. \ref{fig:proj1} and the action pair $\{\hatt_1 = \theta' + \Delta, \hatt_2 = \theta'\}$, where $\theta' \in \Theta \cap [\theta_{d,2}-\delta t/2,\theta_{e,1})$ leads to an equilibrium in Fig. \ref{fig:proj2}. Therefore the corresponding time allocations at equilibria are given by $t_1 = \theta$ and $t_2 = \theta+\delta t + \Delta$ in Fig. \ref{fig:proj1} and given by $t_1 = \theta' +\delta t + \Delta$ and $t_2 = \theta'$ in Fig. \ref{fig:proj2}. However, for the case in Fig. \ref{fig:proj1}, there does not exist a $\theta \in \Theta \cap [\theta_{e,1},\theta_{d,2}-\delta t/2)$  such that $t_1^* = \theta$ and $t_2^* = \theta + \delta t + \Delta$.  Additionally, for the case in Fig. \ref{fig:proj2}, there does not exist a $\theta' \in \Theta \cap [\theta_{d,2}-\delta t/2,\theta_{e,1})$  such that $t_1^* = \theta'  + \delta t + \Delta$ and $t_2^* = \theta'$. Existence of such $\theta,\theta'$ would imply that the socially optimal allocation is in the set of allocations at equilibria. Therefore the socially optimal time allocation may not lead to an equilibrium. However, in such a case, since the set of equilibrium points is finite, there exists at least one equilibrium that is socially more preferable than, or equally preferable with, the other equilibrium points. We call such an equilibrium ``socially optimal equilibrium" and denote the corresponding time allocations by $(t_1^o,t_2^o)$ (e.g., see Figs. \ref{fig:proj1} and \ref{fig:proj2}).

Let $\gamma := \{\theta_{d,1},\theta_{e,1},\theta_{d,2},\theta_{e,2}\}$, and given $\gamma$, let $M_{\gamma} \subset \Theta \times \Theta$ denote the set of all the corresponding equilibrium points and $T_{\gamma} \subset \Theta \times \Theta$ be the corresponding set of allocated passing times.  If there is a unique pure equilibrium, e.g., the case when $\theta_{d,1} = \theta_{d,2}$ and $\theta_{e,1} = \theta_{e,2}$, then this is also the socially optimal equilibrium point. Otherwise, the socially optimal equilibrium leads to the allocations $(t_1^o,t_2^o)$, which are given by
\begin{align}
\min_{(t_1,t_2) \in T_{\gamma}} \sum_{i=1}^2 c(|t_i - \theta_{d,i}|).\label{eq:opt}
\end{align}
The following theorem characterizes these allocations.

\noindent
{\bf Theorem 1.}
{\em Suppose that there are multiple pure equilibria, and yet the socially optimal allocations cannot be achieved at any of these equilibria. If $\theta_{d,1} = \theta_{d,2}$, let $i = \arg\min_{k=1,2} \theta_{e,k}$ and $j = \arg\max_{k=1,2}\theta_{e,k}$; otherwise let $i = \arg\min_{k=1,2} \theta_{d,k}$ and $j = \arg\max_{k=1,2}\theta_{d,k}$. Then, the allocations in the socially optimal equilibrium are given by
\begin{align}\label{eq:allo}
(t_1^o,t_2^o) = \left\{\begin{array}{ll}
(\theta_{e,j},\theta_{e,j} + \delta t + \Delta) & \theta_{d,i} = \theta_{d,j},\\
(\bar{\theta} + \delta t + \Delta, \bar{\theta}) & \hspace{-.05in}\begin{array}{l} \theta_{e,j}<\theta_{e,i}\leq\theta_{d,i}<\theta_{d,j} \\ \mbox{ and } \theta_{d,j}-\delta t/2 < \theta_{e,i}\end{array}\\
(\theta_{e,i}, \theta_{e,i} + \delta t + \Delta) & \mbox{otherwise.}
\end{array}\right.
\end{align}
where $\bar{\theta} := \max\{\theta_{d,j}-\delta t/2,\theta_{e,j}\}$.}

{\em Proof.} The proof is provided in Appendix F. \hfill $\square$

We reemphasize that also the time allocations in the socially optimal equilibrium, $t_1^o,t_2^o$, do not depend on the payoff function directly. The following corollary provides the socially optimal equilibrium points $(\hatt_1^o,\hatt_2^o)$ such that $(t_1^o,t_2^o) = \mathrm{FCFS}(\hatt_1^o,\hatt_2^o)$.

\begin{table*}[t!]
\renewcommand{\arraystretch}{1.2}
\caption{The cases in Corollary 1, where $b \sim \mathrm{Ber}(1/2)$.}
\begin{center}
\begin{tabular}{|c|c|l|l|} \hline
	Case                 & \multicolumn{2}{|c|}{Condition}                                                                                                                                                             & \multicolumn{1}{|c|}{Assignment of $(\hatt_i^o,\hatt_j^o)$ }                         \\\hline\hline
	No Conflict          & \multicolumn{2}{|l|}{$\theta_{d,i} + \delta t < \theta_{d,j}$}                                                                                                                              & $( \theta_{d,i}, \theta_{d,j} )$                                                     \\\hline
	Lemma 1    & \multicolumn{2}{|l|}{$\theta_d := \theta_{d,i} = \theta_{d,j}$}                                                                                                                             &                                                                                      \\\cline{2-4}
	                     & if      & $\max\{\theta_{e,i},\theta_d - \delta t/2\} = \max\{\theta_{e,j},\theta_d - \delta t/2\}$                                                                                         & $( \theta, \theta )$, where $\theta = \max\{\theta_{e,j},\theta_d - \delta t/2\}$    \\
	                     & else if & $\theta^* \in \Theta$                                                                                                                                                             & $( \min\{\theta^*,\theta_{e,j}\}, \hatt_i^o + \Delta )$                              \\
	                     & else if & $\theta_{e,j} \leq \theta^* - \Delta/2$                                                                                                                                           & $( \theta_{e,j}, \hatt_i^o + \Delta )$                                               \\
	                     & else    &                                                                                                                                                                                   & $( b (\theta^* - \Delta/2) + (1-b)(\theta^* + \Delta/2), \hatt_i^o + \Delta )$       \\\hline
	Lemma 2    & \multicolumn{2}{|l|}{$\theta_{e,i},\theta_{e,j} \leq \theta_{d,i} < \theta_{d,j} \leq \theta_{d,i} + \delta t$ and $\theta_{d,j} - \delta t \leq \theta_{d,i} < \theta_{d,j} - \delta t/2$} &                                                                                      \\\cline{2-4}
	                     & if\footnotemark      & $\max\{\theta_{d,j} - \delta t, \theta_{e,i}\} = \theta_{d,i}$                                                                                                                    & $( \theta_{d,i}, \theta_{d,j} )$ \\
	                     & else if & $\theta^* \in \Theta$                                                                                                                                                             & $( \max\{\theta^*,\theta_{e,i}\}, \hatt_i^o + \Delta )$                              \\
	                     & else if & $\theta^* + \Delta/2 \leq \theta_{e,i}$                                                                                                                                           & $( \theta_{e,i}, \hatt_i^o + \Delta )$                                               \\
	                     & else    &                                                                                                                                                                                   & $( b (\theta^* - \Delta/2) + (1-b)(\theta^* + \Delta/2), \hatt_i^o + \Delta )$       \\\hline
	Lemma 3    & \multicolumn{2}{|l|}{$\theta_{e,i} \leq \theta_{e,j} \leq \theta_{d,i} < \theta_{d,j} \leq \theta_{d,i} + \delta t$ and $\theta_{d,j} - \delta t/2 \leq \theta_{d,i}$}                      &                                                                                      \\\cline{2-4}
	                     & if      & $\max\{\theta_{e,i},\theta_{d,i}-\delta t/2\} = \max\{\theta_{d,j}-\delta t/2, \theta_{e,j}\}$                                                                                    & $( \theta, \theta )$, where $\theta = \max\{\theta_{d,j}-\delta t/2, \theta_{e,j}\}$ \\
	                     & else if & $\theta^* \in \Theta$                                                                                                                                                             & $( \max\{\theta^*,\theta_{e,i}\}, \hatt_i^o + \Delta )$                              \\
	                     & else if & $\theta^* + \Delta/2 \leq \theta_{e,i}$                                                                                                                                           & $( \theta_{e,i}, \hatt_i^o + \Delta )$                                               \\
	                     & else    &                                                                                                                                                                                   & $( b (\theta^* - \Delta/2) + (1-b)(\theta^* + \Delta/2), \hatt_i^o + \Delta )$       \\\hline
	Former Case          & \multicolumn{2}{|l|}{$\theta_{e,j} < \theta_{e,i} \leq \theta_{d,i} < \theta_{d,j}$ and $\theta_{e,i} \leq \theta_{d,j} - \delta t/2 \leq \theta_{d,i}$}                                    &                                                                                      \\\cline{2-4}
	in Lemma 4 & if      & $\max\{\theta_{e,i},\theta_{d,i}-\delta t/2\} = \theta_{d,j}-\delta t/2$                                                                                                          & $( \theta, \theta )$, where $\theta = \theta_{d,j}-\delta t/2$                       \\
	                     & else if & $\theta^* \in \Theta$                                                                                                                                                             & $( \max\{\theta^*,\theta_{e,i}\}, \hatt_i^o + \Delta )$                              \\
	                     & else if & $\theta^* + \Delta/2 \leq \theta_{e,i}$                                                                                                                                           & $( \theta_{e,i}, \hatt_i^o + \Delta\} )$                                             \\
	                     & else    &                                                                                                                                                                                   & $( b (\theta^* - \Delta/2) + (1-b)(\theta^* + \Delta/2), \hatt_i^o + \Delta )$       \\\hline
	Latter Case          & \multicolumn{2}{|l|}{$\theta_{e,j} < \theta_{e,i} \leq \theta_{d,i} < \theta_{d,j}$ and $\theta_{d,j} - \delta t/2 < \theta_{e,i} \leq \theta_{d,i}$}                                       &                                                                                      \\\cline{2-4}
	in Lemma 4 & if      & $\max\{\theta_{e,j},\theta_{d,j}-\delta t/2\} = \theta_{e,i}$                                                                                                                     & $( \theta_{e,i}, \theta_{e,i} )$                                                     \\
	                     & else    &                                                                                                                                                                                   & $( \hatt_j^o + \Delta, \min\{\theta_{d,j}-\delta t/2,\theta_{e,j}\} )$               \\\hline
	Lemma 5    & \multicolumn{2}{|l|}{$\theta_{e,i}\leq\theta_{d,i} < \theta_{e,j}\leq\theta_{d,j}\leq \theta_{d,i} + \delta t$}                                                                             &                                                                                      \\\cline{2-4}
	                     & if      & $\max\{\theta_{d,j} - \delta t, \theta_{d,i} - \delta t/2, \theta_{e,i}\} = \min\{\theta_{d,j}-\delta t/2,\theta_{d,i}\}$                                                         & $( \theta, \theta )$, where $\theta = \min\{\theta_{d,j}-\delta t/2,\theta_{d,i}\}$ \\
	                     & else if & $\theta^* \in \Theta$                                                                                                                                                             & $( \max\{\theta^*,\theta_{e,i}\}, \hatt_i^o + \Delta )$                              \\
	                     & else if & $\theta^* + \Delta/2 \leq \theta_{e,i}$                                                                                                                                           & $( \theta_{e,i}, \hatt_i^o + \Delta )$                                               \\
	                     & else    &                                                                                                                                                                                   & $( b (\theta^* - \Delta/2) + (1-b)(\theta^* + \Delta/2), \hatt_i^o + \Delta )$       \\\hline
\end{tabular}\label{table}
\end{center}
\end{table*}

\noindent
{\bf Corollary 1.}
{\em If $\theta_{d,1} = \theta_{d,2}$, let $i = \arg\min_{k=1,2} \theta_{e,k}$ and $j = \arg\max_{k=1,2}\theta_{e,k}$; otherwise let $i = \arg\min_{k=1,2} \theta_{d,k}$ and $j = \arg\max_{k=1,2}\theta_{d,k}$. Let $\theta^* \in \mathbb{R}$ be defined by $\theta^* := \theta_{d,i} - (\delta t - (\theta_{d,j}-\theta_{d,i}))/2$. Then, the socially optimal equilibrium is given in Table \ref{table}.}

{\em Proof.}
These results follow from Theorem 1 and Lemmas 1-5. We select the equilibrium point within $T_{\gamma}$ such that the corresponding time allocation is closest to the socially optimal allocation. \hfill $\square$

Next, the following corollary provides a truthful mechanism for intersection control in strategic environments.

\noindent
{\bf Corollary 2.}
{\em Consider an intersection mechanism, which asks the agents to report both their earliest possible and desired passing times, $\hatt_{e,i},\hatt_{d,i}$, for $i=1,2$; and plays for them the corresponding socially optimal equilibrium actions, i.e., reports passing times to the intersection manager: $\hatt_1^o,\hatt_2^o$, which are formulated in Corollary 1. Then, this intersection mechanism is a strategy-proof mechanism, i.e., it incentivizes the selfish agents to reveal their private information truthfully. }

{\em Proof.}
This follows from the revelation principle \cite{ref:Vazirani07}, which implies that the socially optimal equilibrium can be implemented by an incentive-compatible-direct-mechanism, in which both agents reveal their private information truthfully. The intersection mechanism already plays the best actions strategically for both agents, which lead to the corresponding socially optimal equilibrium; therefore the agents do not have any incentive to misreport their private information with the aim of reducing their own payoff function further. \hfill $\square$
\footnotetext{Note that this is one of the essentially unique equilibria, which lead to the same time allocations. Furthermore, truthfulness of both agents leads to the socially optimal equilibrium.}

\section{CONCLUSION}\label{sec:conclusion}

In this paper, we have studied the truthfulness of intersection control, when the intersection manager evaluates the requests of more than one agent, specifically two agents, in a non-cooperative environment. We analyze equilibria for the agents' actions such that at an equilibrium, given that the other agent has taken that action against his/her action, he/she has no incentive to change his/her action unilaterally. We have formulated all equilibria for all possible scenarios. We have shown that there always exists a pure equilibrium and there even exist multiple pure equilibria in general. We have then characterized the socially preferable equilibrium within all the equilibria with respect to a certain social objective. Finally, we have designed a strategy-proof mechanism, where the agents cannot exploit the intersection control to get benefit in terms of their selfish objectives. We point out that the intelligent intersection control process in a non-cooperative environment even with two agents requires a careful consideration of all the possible cases with respect to earliest and desired passing times of the agents, and the time that the agents need to pass through the intersection. By formulating the strategy-proof mechanism, this paper has also provided a straight-forward, yet non-trivial, guideline to extend to scenarios where there are more than two agents.

\bibliographystyle{IEEEtran}
\bibliography{references}

\begin{thebibliography}{1}
\providecommand{\url}[1]{#1}
\csname url@rmstyle\endcsname
\providecommand{\newblock}{\relax}
\providecommand{\bibinfo}[2]{#2}
\providecommand\BIBentrySTDinterwordspacing{\spaceskip=0pt\relax}
\providecommand\BIBentryALTinterwordstretchfactor{4}
\providecommand\BIBentryALTinterwordspacing{\spaceskip=\fontdimen2\font plus
\BIBentryALTinterwordstretchfactor\fontdimen3\font minus
  \fontdimen4\font\relax}
\providecommand\BIBforeignlanguage[2]{{%
\expandafter\ifx\csname l@#1\endcsname\relax
\typeout{** WARNING: IEEEtran.bst: No hyphenation pattern has been}%
\typeout{** loaded for the language `#1'. Using the pattern for}%
\typeout{** the default language instead.}%
\else
\language=\csname l@#1\endcsname
\fi
#2}}

\bibitem{ref:Dresner08}
K.~Dresner and P.~Stone, ``A multiagent approach to autonomous intersection
  management,'' \emph{Journal of Artificial Intelligence Research}, vol.~31,
  pp. 591--656, 2008.

\bibitem{ref:Vasirani12}
M.~Vasirani and S.~Ossowski, ``A market-inspired approach for intersection
  management in urban road traffic networks,'' \emph{Journal of Artificial
  Intelligence Research}, vol.~43, pp. 621--659, 2012.

\bibitem{ref:Elhenawy15}
M.~Elhenawy, A.~A. Elbery, A.~A. Hassan, and H.~A. Rakha, ``An intersection
  game-theory-based traffic control algorithm in a connected vehicle
  environment,'' in \emph{IEEE International Conference on Intelligent
  Transportation Systems (ITSC)}, Sept. 2015, pp. 343--347.

\bibitem{ref:Han12}
Z.~Han, D.~Niyato, W.~Saad, and T.~Ba\c{s}ar, \emph{Game Theory in Wireless and
  Communication Networks: Theory, Models, and Applications}.\hskip 1em plus
  0.5em minus 0.4em\relax Cambridge, UK: Cambridge University Press, 2012.

\bibitem{ref:Basar99}
T.~Ba\c{s}ar and G.~J. Olsder, \emph{Dynamic Noncooperative Game Theory}.\hskip
  1em plus 0.5em minus 0.4em\relax Philadelphia: SIAM Series in Classics in
  Applied Mathematics, 1999.

\bibitem{ref:Sayin17}
M.~O. Sayin, C.-W. Lin, S.~Shiraishi, and T.~Ba\c{s}ar, ``Information-driven
  intersection management: {T}ruthfulness via payments,'' \emph{Submitted to
  IEEE Transactions on Intelligent Transportation Systems}, 2017.

\bibitem{ref:Vazirani07}
V.~V. Vazirani, N.~Nisan, T.~Roughgarden, and E.~Tardos, \emph{Algorithmic Game
  Theory}.\hskip 1em plus 0.5em minus 0.4em\relax Cambridge, UK: Cambridge
  University Press, 2007.

\end{thebibliography}

\appendices
\section{}
\subsection{Proof of Lemma 1}
Given that $\theta_{d,1} = \theta_{d,2} =: \theta_d$ and $\theta_{e,1} \leq \theta_{e,2}$, we have the following cases, which are partially demonstrated in Fig. \ref{fig:sub1}:
$i)$ If both $\theta_{e,1}$ and $\theta_{e,2}$ are less than or equal to $\theta_{d} - \delta t/2$, there exists a unique pure equilibrium, where both agents report $\hatt_1 = \hatt_2 = \theta_d - \delta t/2$, and the intersection is allocated to one of the agents at $\theta_d - \delta t/2$ with equal probability and the other agent is allowed to pass through the intersection when $\theta_d + \delta t/2 + \Delta$. The action pair $\{\hatt_1 = \theta_d - \delta t/2, \hatt_2 = \theta_d-\delta t/2\}$ leads to an equilibrium since if agent-$1$ reports $\hatt_1 = \theta_d - \delta t/2$, then the best response of agent-$2$ is to report $\hatt_2 = \theta_d - \delta t/2$, and given agent-2 has reported $\hatt_2$, agent-$1$ has no incentive to change his/her action. Furthermore, this is the only pure equilibrium.
$ii)$ If $\theta_{e,2} > \theta_d - \delta t/2$ and $\theta_{e,1} \leq \theta_d - \delta t/2$, then there exist multiple pure equilibria. In particular, if agent-$2$ reports $\hatt_2 = \theta + \Delta$, where $\theta \in \Theta$ such that $\theta \in [\theta_d-\delta t/2,\theta_{e,2})$, then the best response of agent-$1$ is to report $\hatt_1 = \theta$ and given that agent-$1$ has reported that, agent-$2$ has no incentive to change his/her action since agent-$2$ cannot pass through the intersection before $\theta_{e,2}$. Note that the set $[\theta_d - \delta t/2, \theta_{e,2})$ is not empty since $\theta_{e,2} > \theta_d - \delta t/2$.
$iii)$ Finally, if $\theta_{e,2}\geq\theta_{e,1} > \theta_d-\delta t/2$, then there exist multiple pure equilibria. Correspondingly, the equilibrium achieving action pairs are given by $\{\hatt_1 = \theta, \hatt_2 = \theta + \Delta\}$, where $\theta\in\Theta$ such that $\theta \in [\theta_{e,1},\theta_{e,2})$. If there is no such $\theta$, i.e., $\theta_e := \theta_{e,1} = \theta_{e,2}$, then we have a unique pure equilibrium, where both agents report $\hatt_1 = \hatt_2 = \theta_e$.

\subsection{Proof of Lemma 2}
Given that $\theta_{e,1}, \theta_{e,2} \leq \theta_{d,1} < \theta_{d,2} \leq \theta_{d,1} + \delta t$ and $\theta_{d,2} - \delta t \leq \theta_{d,1} < \theta_{d,2}-\delta t/2$, we have the following cases, which are partially demonstrated in Fig. \ref{fig:sub2}:
$i)$ If both $\theta_{e,1}$ and $\theta_{e,2}$ are less than or equal to $\theta_{d,2}-\delta t$, there exist multiple equilibria, where the equilibrium achieving action pairs are given by $\{\hatt_1 = \theta, \hatt_2 = \theta + \Delta\}$, where $\theta \in \Theta$ such that $\theta \in [\theta_{d,2}-\delta t, \theta_{d,1})$. Additionally, the action pair $\{\hatt_1 = \theta_{d,1}, \hatt_2\}$, where $\hatt_2 \in \Theta$ such that $\hatt_2 \in (\theta_{d,1},\theta_{d,1}+\delta t+\Delta]$, also leads to an equilibrium. Note that if agent-$2$ reports a time earlier than $\theta_{d,2}-\delta t$, e.g., $\theta' < \theta_{d,2}-\delta t$, the best response of agent-$1$ is to report a far earlier time, e.g., $\theta' - \Delta$. However, given that agent-$1$ has reported $\theta'-\Delta$, now, agent-$2$ has an incentive to change his/her action since by reporting $\theta_{d,2}$, agent-$2$ can pass through the intersection exactly at his/her desired passing time, which is the least possible payoff he/she can get. Therefore, such action pairs do not lead to an equilibrium.
$ii)$ If $\theta_{d,2} - \delta t \leq \theta_{e,1} \leq \theta_{e,2} \leq \theta_{d,1}$, then there exist multiple equilibria, where the equilibrium achieving action pairs are given by $\{\hatt_1 = \theta, \hatt_2 = \theta + \Delta\}$, where $\theta \in \Theta$ such that $\theta \in [\theta_{e,1},\theta_{d,1})$. Correspondingly,  the action pair $\{\hatt_1 = \theta_{d,1}, \hatt_2\}$, where $\hatt_2 \in \Theta$ such that $\hatt_2 \in (\theta_{d,1},\theta_{d,1}+\delta t+\Delta]$, also leads to an equilibrium.

\subsection{Proof of Lemma 3}
Given that $\theta_{e,1} \leq \theta_{e,2} \leq \theta_{d,1} < \theta_{d,2} \leq \theta_{d,1} + \delta t$ and $\theta_{d,1} \geq \theta_{d,2} -\delta t/2$, we have the following cases, which are partially demonstrated in Fig. \ref{fig:sub3}:
$i)$ If $\theta_{e,1} \leq \theta_{d,1}-\delta t/2$ and $\theta_{e,2} \leq \theta_{d,2} - \delta t/2$, then there exist multiple pure equilibria, where the equilibrium achieving action pairs are given by $\{\hatt_1 = \theta, \hatt_2 = \theta + \Delta\}$, where $\theta \in \Theta$ such that $\theta \in [\theta_{d,1}-\delta t/2, \theta_{d,2}-\delta t/2)$. Note that the set $[\theta_{d,1}-\delta t/2, \theta_{d,2}-\delta t/2)$ is not empty since $\theta_{d,1} > \theta_{d,2}$.
$ii)$ If $\theta_{d,1} - \delta t/2 \leq \theta_{e,1} \leq \theta_{e,2} \leq \theta_{d,2} - \delta t /2$, we have two cases. If $\theta_{e,1} < \theta_{d,2}-\delta t/2$, there exist multiple pure equilibria, where the equilibrium achieving action pairs are given by $\{\hatt_1 = \theta, \hatt_2 = \theta + \Delta\}$, where $\theta \in \Theta$ such that $\theta \in [\theta_{e,1}, \theta_{d,2}-\delta t/2)$. Or if $\theta_{e,1} = \theta_{d,2}-\delta t/2$, there exists a unique pure equilibrium, where $\{\hatt_1 = \theta_{e,1},\hatt_2 = \theta_{e,1}\}$.
$iii)$ If $\theta_{d,1} - \delta t/2 \leq \theta_{e,1} \leq \theta_{d,2} - \delta t/2 < \theta_{e,2}$, there exist multiple pure equilibria, where the equilibrium achieving action pairs are given by $\{\hatt_1 = \theta, \hatt_2 = \theta + \Delta\}$, where $\theta \in \Theta$ such that $\theta \in [\theta_{e,1}, \theta_{e,2})$. Note that the set $[\theta_{e,1},\theta_{e,2})$ is not empty since $\theta_{e,2}>\theta_{e,1}$.
$iv)$ Finally, if $\theta_{d,2} - \delta t/2 < \theta_{e,1} \leq \theta_{e,2}$, we have two cases. If $\theta_{e,1}<\theta_{e,2}$, there exist multiple pure equilibria, where the equilibrium achieving action pairs are given by $\{\hatt_1 = \theta, \hatt_2 = \theta + \Delta\}$, where $\theta \in \Theta$ such that $\theta \in [\theta_{e,1}, \theta_{e,2})$. Or if $\theta_{e,1} = \theta_{e,2}$, there exists a unique pure equilibrium, where $\{\hatt_1 = \theta_{e,1},\hatt_2 = \theta_{e,1}\}$.

\subsection{Proof of Lemma 4}
Given that $\theta_{e,2} < \theta_{e,1} \leq \theta_{d,1} < \theta_{d,2} \leq \theta_{d,1} + \delta t$ and $\theta_{d,1} \geq \theta_{d,2} - \delta t/2$, we have interesting equilibrium scenarios, which are partially demonstrated in Fig. \ref{fig:sub4}, e.g., at equilibrium, agent-$2$ might be allowed to pass first even though his/her desired passing time is later:
$i)$ If $\theta_{e,1} \leq \theta_{d,1} - \delta t/2$, then there exist multiple pure equilibria, where the equilibrium achieving action pairs are given by $\{\hatt_1 = \theta, \hatt_2 = \theta + \Delta\}$, where $\theta \in \Theta$ such that $\theta \in [\theta_{d,1}-\delta t/2, \theta_{d,2}-\delta t/2)$. Note that the set $[\theta_{d,1}-\delta t/2, \theta_{d,2}-\delta t/2)$ is not empty since $\theta_{d,1} > \theta_{d,2}$.
$ii)$ If $\theta_{d,1}-\delta t/2 < \theta_{e,1} \leq \theta_{d,2} - \delta t/2$, we have two cases. If $\theta_{e,1} < \theta_{d,2}-\delta t/2$, there exist multiple equilibria such that the equilibrium achieving action pairs are given by $\{\hatt_1 = \theta, \hatt_2 = \theta + \Delta\}$, where $\theta \in \Theta$ such that $\theta \in [\theta_{e,1},\theta_{d,2}-\delta t/2)$. Or if $\theta_{e,1} = \theta_{d,2} - \delta t/2$, there exists a unique pure equilibrium, where $\{\hatt_1 = \theta_{e,1},\hatt_2 = \theta_{e,1}\}$.
$iii)$ Interestingly, if $\theta_{e,2} \leq \theta_{d,2} - \delta t/2 < \theta_{e,1} \leq \theta_{d,1}$, we have two cases. If $\theta_{d,2} - \delta t/2 < \theta_{e,1}$, multiple equilibria such that the equilibrium achieving action pairs are given by $\{\hatt_1 = \theta+\Delta, \hatt_2 = \theta\}$, where $\theta \in \Theta$ such that $\theta \in [\theta_{d,2}-\delta t/2,\theta_{e,1})$. And at these equilibrium points, agent-$2$ passes through the intersection first. Or if $\theta_{d,2} -\delta t/2 = \theta_{e,1}$, there exists a unique pure equilibrium, where $\{\hatt_1 = \theta_{e,1},\hatt_2 = \theta_{e,1}\}$.
$iv)$ Furthermore, if $\theta_{d,2} - \delta t/2 < \theta_{e,2} < \theta_{e,1} \leq \theta_{d,1}$, then there exist multiple equilibria such that the equilibrium achieving action pairs are given by $\{\hatt_1 = \theta+\Delta, \hatt_2 = \theta\}$, where $\theta \in \Theta$ such that $\theta \in [\theta_{e,2},\theta_{e,1})$. And again at these equilibrium points, agent-$2$ passes through the intersection first.

\subsection{Proof of Lemma 5}
Finally, we consider the cases that $\theta_{e,1} \leq \theta_{d,1} < \theta_{e,2} \leq \theta_{d,2} \leq \theta_{d,1}+\delta t$, which are partially represented in Fig. \ref{fig:sub5}.
$i)$ If $\theta_{e,1} \leq \theta_{d,2} - \delta t \leq \theta_{d,1} \leq \theta_{d,2}-\delta t/2$, then we have two cases. If $\theta_{d,2}-\delta t/2 < \theta_{d,1}$, there exist multiple equilibria such that the equilibrium achieving action pairs are given by $\{\hatt_1 = \theta, \hatt_2 = \theta + \Delta\}$, where $\theta \in \Theta$ such that $\theta \in [\theta_{d,2}-\delta t/2,\theta_{d,1})$. Or if $\theta_{d,2} - \delta t/2 = \theta_{d,1}$, there exists a unique pure equilibrium, where $\{\hatt_1 = \theta_{d,1},\hatt_2 = \theta_{d,1}\}$.
$ii)$ If $\theta_{d,2} - \delta t < \theta_{e,1} \leq \theta_{d,1} \leq \theta_{d,2}-\delta t/2$, then we have two cases. If $ \theta_{e,1} < \theta_{d,1}$, there exist multiple equilibria such that the equilibrium achieving action pairs are given by $\{\hatt_1 = \theta, \hatt_2 = \theta + \Delta\}$, where $\theta \in \Theta$ such that $\theta \in [\theta_{e,1},\theta_{d,1})$. Or if $\theta_{d,2} - \delta t/2 = \theta_{d,1}$, there exists a unique pure equilibrium, where $\{\hatt_1 = \theta_{d,1},\hatt_2 = \theta_{d,1}\}$.
$iii)$ If $\theta_{d,2} - \delta t/2 < \theta_{d,1}$, then there exist multiple equilibria such that the equilibrium achieving action pairs are given by $\{\hatt_1 = \theta, \hatt_2 = \theta + \Delta\}$, where $\theta \in \Theta$ such that $\theta \in [\theta_{d,1}-\delta t/2,\theta_{d,2}-\delta t/2)$. Note also that the set $[\theta_{d,1}-\delta t/2, \theta_{d,2}-\delta t/2)$ is not empty since $\theta_{d,1} > \theta_{d,2}$.

\begin{figure}[t!]
\centering
\includegraphics[width = .4\textwidth]{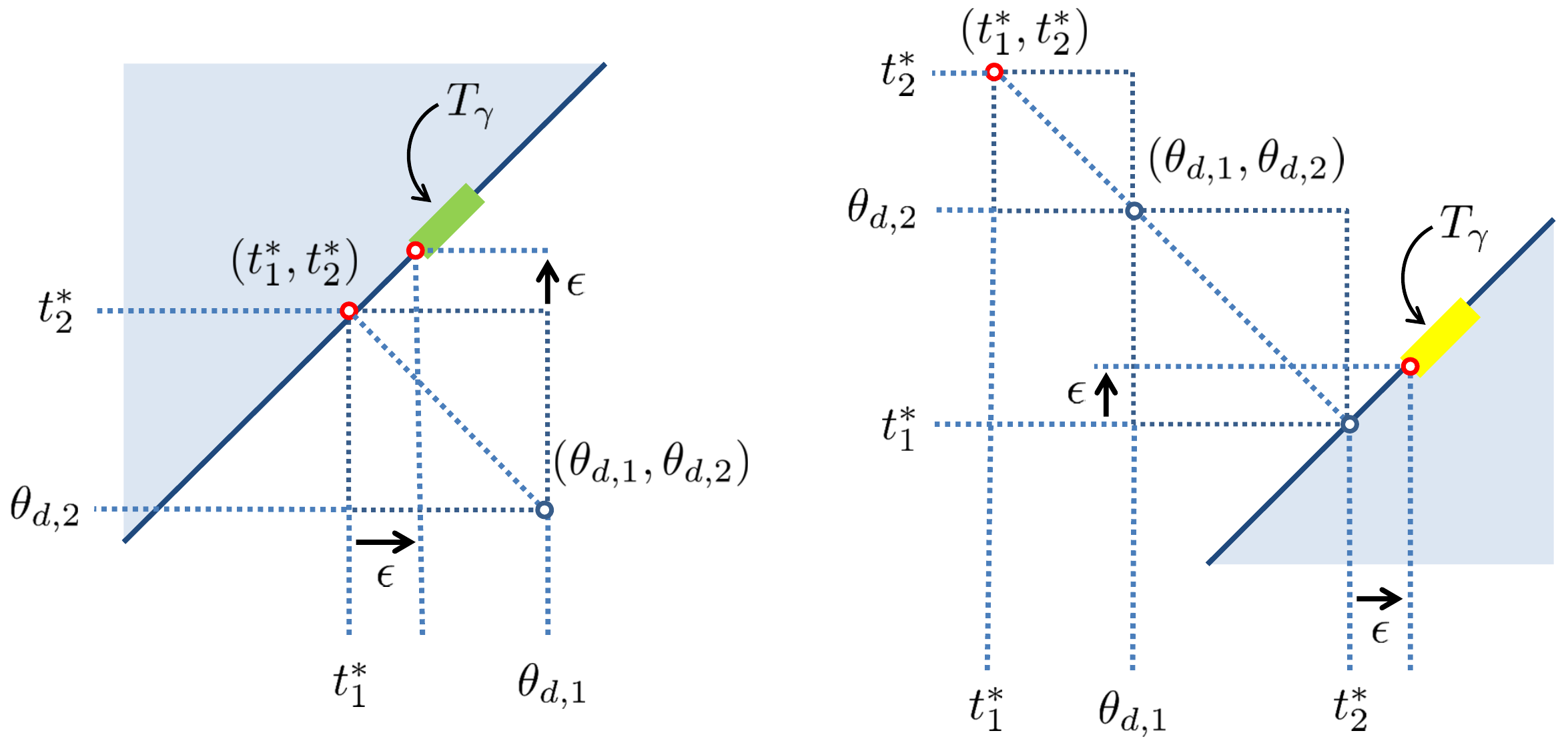}
\caption{The distance between the allocations in socially optimal equilibrium and the socially optimal allocations in distinct equilibrium scenarios, where both agents can pass through the intersection first (on the left figure, agent-$1$ passes first while on the right figure agent-$2$ passes first) in spite of $\theta_{d,1} < \theta_{d,2}$.} \label{fig:proj3}
\end{figure}

\subsection{Proof of Theorem 1}
We seek to solve \eqref{eq:opt} in order to formulate the allocations at the socially optimal equilibrium. To this end, we provide the illustrations in Fig. \ref{fig:proj3}. first, consider the illustration on the left of Fig. \ref{fig:proj3}. Here, all the allocation pairs in $T_{\gamma}$ can be written as $(t_1^* + \epsilon, t_2^* + \epsilon)$ for certain $\epsilon \in \mathbb{R}$. Note that $\epsilon < 0$ would imply $T_{\gamma}$ stands on the left of $(t_1^*,t_2^*)$, i.e., such cases are also considered. Then, the optimization problem \eqref{eq:opt} can be written as
\begin{equation}\label{eq:opt2}
\min_{\epsilon \in \mathbb{R} : (t_1^* + \epsilon,t_2^* + \epsilon) \in T_{\gamma}} c(|t_1^* + \epsilon - \theta_{d,1}|) + c(|t_2^* + \epsilon - \theta_{d,2}|).
\end{equation}
Since at equilibria, agent-$1$ would not pass through the intersection after $\theta_{d,1}$ or before $\theta_{d,1} - \delta t/2$ while he/she is passing first, \eqref{eq:opt2} can also be written as
\begin{equation}
\min_{|\epsilon| : (t_1^* + \epsilon,t_2^* + \epsilon) \in T_{\gamma}} c(\sigma_1 - \epsilon) + c(\sigma _1+ \epsilon),
\end{equation}
where $\sigma_1 := \theta_{d,1} - t_1^* = t_2^* - \theta_{d,2} > 0$. This would imply that the minimizer $\epsilon^* = \min \{\epsilon \in \mathbb{R} : (t_1^* + \epsilon,t_2^* + \epsilon)\in T_{\gamma}\}$ since $c(\cdot)$ is a strictly increasing strictly convex function over $[0,\infty)$. In particular, in these scenarios, the pair of allocations $(t_1^o,t_2^o)$ in socially optimal equilibrium is a point in $T_{\gamma}$, which is closest to the socially optimal allocation $(t_1^*,t_2^*)$.

Next, consider the illustration on the right of Fig. \ref{fig:proj3}. Here, all the allocation pairs in $T_{\gamma}$ can be written as $(t_2^* + \epsilon, t_1^* + \epsilon)$ for certain $\epsilon \in \mathbb{R}$. We emphasize the difference from the previous case. Then, the optimization problem \eqref{eq:opt} can be written as
\begin{equation}\label{eq:opt3}
\min_{|\epsilon| : (t_2^* + \epsilon,t_1^* + \epsilon) \in T_{\gamma}} c(|t_2^* + \epsilon - \theta_{d,1}|) + c(|t_1^* + \epsilon - \theta_{d,2}|).
\end{equation}
Since at equilibria, agent-$2$ would not pass through the intersection after $\theta_{d,2}$ or before $\theta_{d,2} - \delta t/2$ while he/she is passing first, \eqref{eq:opt3} can also be written as
\begin{equation}
\min_{\epsilon \in \mathbb{R} : (t_2^* + \epsilon,t_1^* + \epsilon) \in T_{\gamma}} c(\sigma_2 + \epsilon) + c(\sigma_2 - \epsilon),
\end{equation}
where $\sigma_2 := \theta_{d,2} - t_1^* = t_2^* - \theta_{d,1} > 0$. Correspondingly, this would imply that the minimizer $\epsilon^* = \min \{\epsilon \in \mathbb{R} : (t_2^* + \epsilon,t_1^* + \epsilon)\in T_{\gamma}\}$ since $c(\cdot)$ is a strictly increasing strictly convex function over $[0,\infty)$. Note that the pair of allocations $(t_1^o,t_2^o)$ in socially optimal equilibrium is also a point in $T_{\gamma}$, which is closest to the socially optimal allocation $(t_1^*,t_2^*)$ in addition to $(t_2^*,t_1^*)$. This also implies that the pair of allocations $(t_1^o,t_2^o)$ in socially optimal equilibrium when $\theta_{d,1} = \theta_{d,2}$ is also a point in $T_{\gamma}$, which is closest to the socially optimal allocation $(t_1^*,t_2^*)$. Hence, under the conditions in Theorem 1, we can conclude that the pair of allocations $(t_1^o,t_2^o)$ in socially optimal equilibrium is a point in $T_{\gamma}$, which is ``closest" to the socially optimal allocation $(t_1^*,t_2^*)$. Based on this conclusion and Lemmas 1-5, the allocations that are in socially optimal equilibrium are given by \eqref{eq:allo}.

\end{document}